\documentclass[preprint,12pt]{elsarticle}

\usepackage{graphicx}
\usepackage{amssymb}
\usepackage{comment}
\usepackage{algorithm}
\usepackage{algcompatible}

\usepackage{algpseudocode}
\usepackage{algorithmicx}

\usepackage{lineno}

\journal{Journal Name}

\begin{document}

\begin{frontmatter}


\title {Building an Effective Intrusion Detection System using Unsupervised Feature Selection in Multi-objective Optimization Framework}


\author[1]{Chanchal Suman}
\author[1]{Somanath Tripathy}
\author[1]{Sriparna Saha}
\address[1]{Computer Science and Engineering Department, Indian Institute of Technology Patna, India.}



\begin{abstract}
Intrusion Detection Systems (IDS) are developed to protect the network by detecting the attack. The current paper proposes an unsupervised feature selection technique for analyzing the network data. The search capability of the non-dominated sorting genetic algorithm (NSGA-II)  has been employed for optimizing three different objective functions utilizing different information theoretic measures including mutual information, standard deviation, and information gain to identify mutually exclusive and a high variant subset of features. Finally, the  Pareto optimal front of the different optimal feature subsets are obtained and these feature subsets are utilized for developing classification systems using different popular machine learning models like support vector machines, decision trees and k-nearest neighbour (k=5) classifier etc.  We have evaluated the results of the algorithm on KDD-99, NSL-KDD and Kyoto 2006+ datasets. The experimental results on  KDD-99 dataset show that decision tree provides better results than other available classifiers. The proposed system obtains the best results of 99.78\% accuracy,  99.27\% detection rate and false alarm rate of 0.2\%,  which are better than all the previous results for KDD dataset. We achieved an accuracy of 99.83\% for 20\% testing data of NSL-KDD dataset  and 99.65\% accuracy for 10-fold cross-validation on  Kyoto dataset.
The most attractive characteristic of the proposed scheme is that during the selection of appropriate feature subset, no labeled information is utilized and different feature quality measures are optimized simultaneously using the multi-objective optimization framework. 
\end{abstract}
\begin{keyword}
Feature selection  \sep Intrusion detection system  \sep Machine learning  \sep Multi-objective optimization  \sep  Non-dominated sorting genetic algorithm
\end{keyword}


\end{frontmatter}


\section{Introduction}
Due to the massive growth of Internet usage and a huge amount
of online data, it is essential to take care of the network. 
Traditional security techniques such as data encryption, user authentication, firewall, etc. are not sufficient to provide trusted security to the network, as technologies are expanding day by day. Intruders are getting different ways of network attacks, so we must have to go for the second line of defense such as Intrusion detection system \cite{ambusaidi2016building}. Intrusion is the set of actions which attempt to harm the Integrity, Confidentiality and Availability of a system. An intrusion detection system is a primary tool which is used for protecting networks and information systems against the threats. It monitors the host or packets transmitted throughout the network. If it detects a security policy violation, then it raises alarm to the system administrator. \par  


The network generates large traffic and this huge data slows down the process of intrusion detection. The data also contains some information which is irrelevant and redundant for the detection purpose, so it is very important to select only those information which is relevant. Thus, feature selection is an important component of an IDS, which can powerfully identify a subset of most relevant features within a dataset to decrease the time for computation.  Features extracted from IDS dataset contain similar types of information and possess high degrees of associations or correlations. Thus deletion of some of these features do not decrease the classification power of the system. The use of a full set of features increases the complexity of the system as well as decreases the accuracy. 
So the selection of a proper subset of features which are highly relevant for the given task as well as uncorrelated to each other is desired \cite{chebrolu2005feature}. 
 IDS developed in \cite{peddabachigari2007modeling, ambusaidi2015unsupervised,thaseen2017intrusion,  salunkhe2017security, wang2018abstracting, aljawarneh2018anomaly, wang2010new, horng2011novel} used  different feature deduction methods for selecting relevant features, whereas IDS presented in  \cite{benaicha2014intrusion, gharaee2016new, aghdam2016feature, dhopte2014genetic, eesa2015novel, shah2018efficient} used evolution theory to detect good subset of features and achieve good accuracy.  Some researchers also used deep learning based techniques for designing IDS \cite{najizad2017optimization,kim2016long,kim2017effective,javaid2016deep, putchala2017deep, chawla2017deep}. 


In all these works, researchers have applied a single optimization function for the process of feature selection. Features can be redundant and correlated simultaneously. We should select the features in such a manner so that there would be minimum redundancy and correlation. The domain of the feature also defines its importance. If there is a feature having the same value for all the samples, then it would not be helpful for the classification, at the same time if it will have a vast range of values for the feature then also it would not be beneficial. Thus we need to identify the feature set in such a way that all these properties should be preserved so that it will be helpful for getting maximum accuracy in minimum computation time.

\par
Motivated by these facts, in the current work, we have devised an effective IDS framework based on an unsupervised feature selection technique. Our IDS framework is divided into two phases. The first phase builds on the search capability of a popular multi-objective optimization based technique, namely non-dominated sorting genetic algorithm-II, NSGA-II\cite{deb2000fast} for optimizing multiple feature quality measures simultaneously, in order to find optimal feature subsets. We have used three different feature quality measures, two of them help in getting only relevant features and removing redundancy from the dataset with the help of some similarity measures ( such as mutual information, Pearson correlation coefficient etc.). Last one helps in getting those features whose variances are high over the samples. Note that the proposed feature selection technique is fully unsupervised. It does not utilize any labeled data at the time of feature selection. The second phase uses different available machine learning classifiers for generating an effective IDS. The features extracted from the first phase can be utilized in different machine learning based classifiers. The key contributions of this paper are as follows:
\begin{itemize}

\item[1.]  All the features present in the IDS data set are not suitable for classifying the data; thus we have devised an unsupervised based framework for evaluating the feature subset. It does not utilize any labeled information during its computation for selecting the relevant feature subset from any IDS data set. 

\item[2.] We have used multi-objective optimization (MOO) for selecting suitable feature subsets. 
Different feature quality measures are optimized simultaneously using the search capability of MOO. Experiments are conducted varying the feature quality measures ranging from mutual information, standard deviation,  information gain, Pearson correlation coefficient, entropy etc.  The qualities of the selected feature subsets are further verified using different machine learning based classifiers. Several IDS systems are developed varying the classification technique and feature subset.  For all the cases good performance is obtained. 

\item[3.] A maximum accuracy  of 99.78 \% is attained by decision tree based IDS in multi-class classification with the use of feature subset identified by MOO-based approach with information gain and standard deviation as objective functions. To the best of our knowledge, this is the best-reported accuracy compared to all the existing works available in the literature. 
\end{itemize}
The rest of this paper is arranged as follows: in section \ref{sec:Related}, we have discussed some prior works which have been done by researchers. In section \ref{background}, important concepts are discussed. In section \ref{MOO_framework}, the framework designed for proposed IDS is discussed. In Section \ref{Proposed_Approach}, the proposed approach for MOO-based feature selection is discussed. the functioning of the algorithm varying the base performance metric is reported. In section \ref{Results}, different simulation results are discussed. In section \ref{results_final}, final results of the proposed IDS are elucidated. Finally, we conclude the paper in section \ref{Conclusion and future work}.

\section{Related Works} \label{sec:Related}
In recent years, some feature selection approaches using different machine learning algorithms are devised  to increase the accuracy and to reduce overhead complexities. Researchers developed a feature selection method using mutual information and Pearson correlation coefficient measure for designing an effective IDS \cite{ambusaidi2016building}. In \cite{chebrolu2005feature}, authors have proposed an ensemble based decision tree classifier for intrusion detection system. They have used Hidden Markov Model (HMM) for finding the feature subsets. In \cite{peddabachigari2007modeling},  authors have proposed decision tree (DT) and support vector machine (SVM) based intrusion-detection models. 
In \cite{ambusaidi2015unsupervised}, authors have devised an unsupervised feature selection algorithm using Laplacian score.  In \cite{thaseen2017intrusion}, authors used chi-square method for feature selection and multi-class support vector machine
(SVM). In this work, Gamma and the fitting constant value for the Radial Basis function are also optimized, to get the good classification results.  Researchers have also used an ensemble of different classifiers in order to increase the accuracy of system  \cite{salunkhe2017security}. In \cite{wang2018abstracting}, authors proposed three different strategies to extract relevant features. After feature extraction, they applied Support vector machine for detection.   In \cite{aljawarneh2018anomaly}, authors built a hybrid model made of J48, Meta Pagging, RandomTree, REPTree, AdaBoostM1,
DecisionStump and NaiveBayes classifier, which resulted a very good accuracy.  In \cite{wang2010new}, researchers combined artificial neural network with fuzzy clustering to achieve better accuracy.  In \cite{horng2011novel}, hierarchical clustering is used for feature selection and then support vector machine is used for classification. 

In \cite{benaicha2014intrusion},  authors proposed an effective IDS using Genetic algorithm. They have used a weighted sum of support and confidence values as the evaluation function. In 2016, authors used the sum of true positive rate, false positive rate and number of selected features to evaluate the fitness of different feature combinations using Genetic algorithm \cite{gharaee2016new}.  Authors have also used Ant-colony optimization based method to find the best feature subset for the Intrusion detection system \cite{aghdam2016feature}.   In \cite{dhopte2014genetic}, researchers have used a combination of a number of connections in the DARPA dataset as fitness function and applied the genetic algorithm for generation of rules to classify the instances. In \cite{eesa2015novel}, authors have used the cuttlefish algorithm as a search technique to find the optimal subset of features. After finding the optimal feature subset, decision tree classifier
was used for checking the performance of the selected features produced by their algorithm. In 2018, authors used Genetic algorithm and support vector machine for classification. GA is
used to optimize all the parameters of SVM, and then SVM is used for efficient intrusion detection \cite{shah2018efficient}.

   In \cite{najizad2017optimization}, authors used artificial neural network to increase the accuracy of the system. In \cite{kim2016long}, authors proposed an IDS using LSTM and efficiency of the technique is improved using different optimizers with LSTM in \cite{kim2017effective}. In \cite{javaid2016deep}, authors used auto encoder for developing IDS. They have used the self-taught learning capacity of autoencoder to learn the features so that good classification results can be achieved. Authors also used gated recurrent neural network for designing an effective IDS \cite{putchala2017deep}. In \cite{chawla2017deep}, the author has designed a real-time Intrusion detection module for extracting features from the network and then applied a sequential neural network having three hidden layers to detect the attack.

\section{Background}  \label{background}
 In this section, We have discussed in brief the problem formulation of feature selection, NSGA-II\cite{deb2000fast},  different criteria for selecting features, different data-sets used for designing the system, and different performance measures. We used the NSGA-II  algorithm for selecting relevant features from the huge dataset, to increase the classification accuracy as well as to decrease the time complexity of the IDS.

\subsection{Feature selection} \label{feature_selection}
In a complex classification system, some features contain false correlations and result in hindering the classification process. Some features may be redundant also. Extra features can lead to an increase in computation time and may impact on the accuracy of the detection system. Feature selection improves
classification by searching for the subset of features, which best classify the training data \cite{chebrolu2005feature}. Thus automatic selection of attributes in training and test data can help in developing a best predictive model.
The objectives of feature selection are: (i) to improve the prediction performance of the model, (ii) to provide a faster and cost-effective model and (iii) to provide a better understanding of the process that generated the data \cite{guyon2003introduction}. Researchers devised many new algorithms for selecting only relevant features from the huge KDD-99 dataset to increase the accuracy of the IDS, which have been already discussed in section \ref{sec:Related}  .  

\subsection{Multi-objective Optimization}
Multi-objective optimization is a technique for optimizing more than one objective functions simultaneously. A popular multi-objective optimization algorithm is NSGA-II (non-dominated sorting genetic algorithm-II) which  is a fast and elitist technique   for optimizing different objective functions. It was proposed in \cite{deb2000fast}. The optimization of multiple objectives associated to a problem leads to the generation of a set of optimal solutions known as Pareto-optimal solutions, instead of a single solution. As none of these solutions is said to be better than the other, we would have many Pareto-optimal solutions for a problem. Below we briefly describe the steps of NSGA-II. \par
NSGA-II is a variant of genetic algorithm (GA) \cite{oh2004hybrid}. Below we have discussed the steps of GA first.  Genetic algorithm is an optimization and search methodology. It uses a chromosome-like data structure for representing solutions which are evolved using selection, recombination and mutation operators.  In GA, chromosomes are represented as linear strings of symbols.  
For the feature selection problem, generally binary encoding is used. If the feature set contains $n$ number of features, then  a string with $n$ binary
digits is used. Each binary digit represents a feature, value of 1 is used to represent the selection of the feature and 0 is used to represent the rejection of the feature. \par
  It works with a set of candidate solutions called as population.  The chromosomes of the populations are  randomly generated binary strings. Each bit of the string is initialized by ''randint(0,1)" function of python denoting the presence or absence of a single feature.  It obtains the optimal solution after a series of iterative computations.  Chromosomes are selected by evaluating the fitness values. A fitness function  is an evaluation function which assesses the quality of a chromosome in every step of evaluation. Selection, crossover and mutation are three evolutionary operators which are repeated in the sequence until termination condition is satisfied. Selection selects the best solution from the population. Cross-over does the job of recombination, and mutation adds some changes in the new population. In this way, using the search capability of GA optimal solution is found \cite{oh2004hybrid}. \par

In  NSGA-II, the number of optimal solutions can not be one, because of the optimization of more than one objective functions. Thus, in NSGA-II a random population $P$ is generated as an initial population and it is sorted based on non-domination.  Using selection, cross-over,  and mutation operators, off-springs are generated. Parents and off-springs are combined and then for the combined population, fronts are determined. The approaches of non-dominated sorting and front calculation are described below in Section \ref{fronts}. Further, according to crowding distance comparison operator, (discussed in \ref{crowd}) the best solutions are kept in the population for the next generation. In this way,  NSGA-II  optimizes multiple objectives as well as keeps the best solutions in the Pareto optimal front. 

\subsubsection{Fast Non-dominated sorting approach} \label{fronts}
In this approach, the set of solutions which are not dominated by any other solution are determined. For each solution two entities: 1) domination count $n_x$- the number of solutions which dominate the solution x, and 2) $S_x$- a set of solutions which are dominated by solution x are calculated. Let there be two solutions, x and y. If a solution x has a better value in at least one of the objective functions and not poor values in other objective functions with respect to y, then y is said to be dominated by x; hence $n_y$ increases by 1  and y is added in the $S_x$.  All solutions in the first non-dominated front will have their domination counts as zero. The front is the set of solutions. For each solution x with  $n_x$ =0, each member y of its set $S_x$ is visited and then $n_y$  is reduced by 1. In this process, if for any member $n_y$ becomes zero, it is put in another list called as second non-dominated front. This procedure is continued until all fronts are identified.
\subsubsection{Crowding-distance calculation and crowded comparison operator} \label{crowd}
After finding all the fronts, the density of solutions surrounding a particular solution in the population needs to be calculated. It is done by calculating the average distance of two solutions on either side of the particular solution along each of the objectives. This is called crowding distance. The sum of individual distance values corresponding to different objective functions is called the overall crowding-distance value. Each objective function is normalized before calculating the crowding distance.The crowded-comparison operator guides the selection process to form a uniformly spread-out Pareto optimal front. 
Every solution in the population has two attributes: non-domination rank (front no.) and crowding distance. Between two solutions with different non domination ranks, the solution with the lower rank is preferred. If both solutions belong to the same front, then a less crowded solution is preferred.


\subsection{Different criteria for selecting features}
There are different criteria which state about the nature of a feature in a dataset. These criteria reveal how useful a feature can be for the classification task without using the class labels. Some such criteria are mutual information, Pearson correlation coefficient, information gain,  entropy and standard deviation.    
\subsubsection{Entropy}
Entropy measures the impurity of a feature. Higher the entropy, more information the feature will have; it means more it will help in predicting the class labels. The entropy of a discrete variable Y can be calculated as :
\begin{equation}
H(Y)= \sum_{y \epsilon Y}^{}-P_ylogP_y
\end{equation}
Where $P_y$ denotes the probability mass function of Y.

\subsubsection{Information gain (IG)} 

Information gain (IG) \cite{wiki:IG}  measures how much information a feature provides us about the class. It measures the change in entropy after using the attribute. Thus, it conveys how important a given feature is. It is calculated as follows:
\begin{equation}
IG = Entropy(parent)- Avg[Entropy(children)] 
\end{equation}
\subsubsection{Mutual information (MI) }
 Mutual information  \cite{goshtasby2012similarity} is  the amount of information communicated in one random variable
about another. "An important theorem from information theory says that the mutual information between two variables is 0 if and only if the two variables are statistically independent". \\
Given two continuous random variables X=\big\{$x_1$, $x_2$,.., $x_n$\big\} and Y=\big\{ $y_1$, $y_2$,..,$y_n$\big\} where n is the total number of
samples, the mutual information between X and Y is
defined as:

\begin{equation}
I(X:Y)=\sum_{x \epsilon X} \sum_{y \epsilon Y} p(x,y) log_2 \frac{p(x,y)}{p(x).p(y)}
\end{equation}
Where, $p(x)$ and $p(y)$ are marginal probability distributions of $x$ and $y$, respectively.
\subsubsection{Pearson correlation coefficient(PCC)} Pearson correlation coefficient \cite{goshtasby2012similarity} measures the linear correlation between two random features. It is symmetric in nature. The value of PCC falls in a definitely closed interval [-1,1]. PCC value close to either -1 or 1 indicates the strong relationship between the two variables. PCC value close to 0 infers
the weaker relationship between them. PCC value 0 indicates no relationship between them. 
PCC quantifies the degree to which a relationship between two variables can be described by a single line.
\begin{equation}
\rho (x,y)= \frac{\sum (x-\bar{x})(y-\bar{y})}{\sqrt{\sum((x-\bar{x})^2.(y-\bar{y})^2)}}
\end{equation}
Where, $\bar{x}$ = Mean of $x$ variable, and $\bar{y}$ = Mean of $y$ variable.

\subsubsection{Standard deviation}
Standard deviation is a measure, which is used to quantify how values for a feature are deviated from the average. It is calculated as follows:
\begin{equation}
SD(\sigma)= \sqrt{\frac{1}{N}\sum_{i=1}^{N}(x_i- \mu)^2}
\end{equation}
Where: \
$SD(\sigma)$ = standard deviation of the feature \newline
$N$ = no. of samples in the feature \newline
$x_i$ = value of ith sample in feature, and \newline
$\mu$ = mean value of the feature

\subsection{Datasets used} \label{datasets}
In order to show the effectiveness of the proposed
approach, we have used the standard  KDD-99 dataset, 
NSL-KDD dataset, and Kyoto 2006+ dataset.
\subsubsection{KDD Cup 99}
The KDD Cup 99  dataset was derived from the
DARPA 98 dataset generated from the 1998 DARPA Intrusion Detection Evaluation program \cite{tavallaee2009detailed}.  It has more than 4 million training samples and 3 million test samples. It contains TCP connection records having 41 informational features plus
one labeled feature. The recorded details of each TCP connection are described by the informational features and the labeled feature specifies the type of connection. By the connection type, it means whether a connection is normal or abnormal. 
The 41 features consist of 32 continuous features and 9 nominal features. The features are classified into 4 categories: basic features, content-based feature, time-based traffic feature, and host-based traffic feature. We have used ''kddcup.data 10 percent” as training data and ''Corrected” as testing data. We have used 10 fold cross validation for validation purpose.
In table \ref{List of different attacks}, we have tabulated different attack classes of training data, and the category they belong to. In the testing data, some more attacks are present, but we considered only those attack classes which come under DoS, Probe, U2R, and R2L.
\begin{table}
\begin{center}
\caption{List of different attacks belonging to different categories }
\begin{tabular}{|p{0.2\linewidth}|p{0.6\linewidth} |}
\hline Attack category & List of attacks \\
\hline DoS & back, neptune, land, pod, smurf, teardrop \\
\hline Probe & ipsweep, nmap, portsweep, satan \\
\hline U2R & buffer\_overflow, loadmodule, rootkit, perl\\
\hline R2L & ftp\_write, guess\_passwd, imap, multihop, phf, spy, warezclient, warezmaster\\
\hline
\end{tabular}

\label{List of different attacks}

\end{center}
\end{table}

\begin{table}
\caption{Data distribution per class in training and testing sets of KDD-99 datset }
\begin{tabular}{|p{0.1\linewidth}|p{0.1\linewidth}|p{0.1\linewidth}|p{0.1\linewidth}|p{0.1\linewidth}|p{0.1\linewidth}| p{0.1\linewidth}|}
\hline data & Dos & Probe & U2R & R2L & Normal \\
\hline train data & 391458 & 4107 & 52 &  1126 & 97278 \\
\hline test data & 222200 & 2377 & 39 & 5993 & 60593 \\
\hline
\end{tabular}
\label{Data}
\end{table}

\subsubsection{NSL-KDD}

Although the KDD Cup 99 dataset is the most widely used benchmark in intrusion detection research, the dataset has some drawbacks. There are many duplicate records, which cause a biased training of classifier. The level of difficulty of the KDD cup 99 dataset is also not very good.
 Thus, to eliminate these undesirable qualities of KDD Cup 99 dataset,
authors of \cite{tavallaee2009detailed} proposed a  more effective dataset, "NSL-KDD dataset". It is based on KDD Cup 99 dataset.  The redundant records of the KDD dataset have been eliminated and the structure of the dataset is reconstructed to increase the level of difficulty. The elimination and reconstruction have made the new dataset more reasonable in both data structure and data size wise.
Therefore, the "NSL-KDD dataset" can be considered as a more standard
dataset for intrusion detection research. We have used the "NSL-KDDTrain+20\%" dataset for the experimental purpose. It is made up of 25192 instances, in which 13449 are normal data and 11743 are considered as attack data. 

\subsubsection{Kyoto 2006+} \label{Kyoto}
The Kyoto 2006+ dataset was presented by \cite{song2006description}. The data was collected over the period from  August 2009 to  November 2009. It was collected from honeypots and regular servers, deployed at Kyoto University. Each connection in this dataset has 24 different features. First 14 are same as KDD-99 and 10 additional features are also present in the data. The additional 10 features enable us to investigate more effectively what happened to our networks. The additional features include: IDS\_detection, Malware\_detection, label etc. The labels are " 1, -1, and -2", where 1 means the session was observed, -1 means attack was observed, and -2 means an unknown attack was observed. 

\subsection{Performance Measures} \label{perform}

Traditionally, researchers use accuracy, detection rate, precision  and  false alarm rate \cite{elhamahmy2010new}  to evaluate the performance of an IDS. The confusion matrix is a tabular structure representing the predicted/ actual classification. It leads to the calculation of True Positive(TP), True negative(TN), False positive(FP) and False negative(FN).  True positive is the number of actual attacks classified as the actual attack. A true negative is the number of non-attacks classified as non-attack. False positive is the number of non-attacks classified as attack class and false negative is the number of actual attacks classified as non-attack. Below we have listed the formulas of different metrics.\\ 
True positive rate or Recall or Detection rate: It is the proportion of test results which are correctly classified by the model.
\begin{equation}
DR = \frac{TP}{(TP+FN)}
\end{equation}
Precision or positive predictive value : It is the probability of correct classification of any instance by the model.
\begin{equation}
Precision=\frac{TP}{(TP+FP)}
\end{equation}
Fall out or false positive rate : It indicates that an attack is predicted by the model, while in reality it does not exist.
\begin{equation}
FPR=\frac{FP}{(FP+TN )}
\end{equation}
Overall accuracy : It is the ratio of number of correctly classified instances and total number of instances.
\begin{equation} \label{accuracy}
ACC=\frac{(TP+TN)}{(TP+FP+FN+TN)}
\end{equation}
Weighted Average Accuracy: The average accuracy (\ref{accuracy}) might not be a good measure of performance in case of imbalance data. Weighted average accuracy is calculated by dividing the sum of product of the accuracy achieved in different classes with its corresponding number of samples and the total number of samples present in the data.
\begin{equation} \label{wt_accuracy}
Weighted-Average-Accuracy  = \frac{\sum_{i=1}^{n}{N_i * Acc_i}}{\sum_{i=1}^{n}{N_i}}
\end{equation}

Where, $N_i$ is the number of samples in ith class, $Acc_i$ is the accuracy of the ith class, and $n$ is the total no. of different classes. \\
F-measure is the harmonic mean of precision and recall. It is used to examine accuracy of a classification system by considering both precision and recall.  
\begin{equation}
F-measure= \frac{2(Precision*Recall)}{(Precision+Recall)}
\end{equation}

\begin{figure*}
 \centering
 \includegraphics[width=1.0\linewidth]{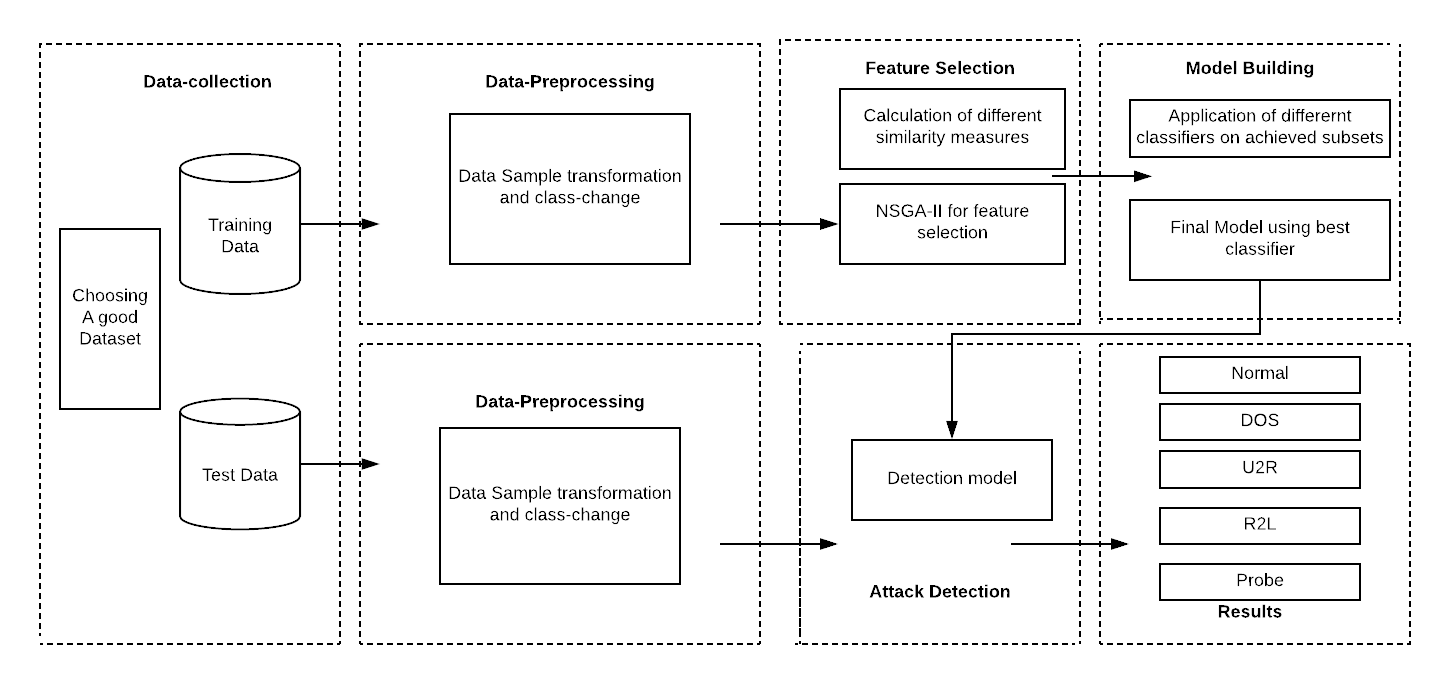}
  \caption{The Proposed Architecture}
  \label{fig:architecture}%
\end{figure*}

\section{Multi Objective Optimization Based Intrusion 
Detection Framework} \label{MOO_framework}
The framework of the proposed IDS
is depicted in \ref{fig:architecture}. It is divided into four main phases. 1) Data collection: In this phase, a good dataset is chosen in order to design the proposed model and evaluate its performance.  (2) Data preprocessing: In this phase, training and test data are preprocessed and normalized, (3) Feature Selection: Different feature quality  measures are computed and a MOO based feature selection technique is employed for determining the Pareto optimal feature subsets, and (4) Model building: Different classifiers are applied on the determined feature subsets to identify the best feature subset and the best classifier. 
For test data, firstly preprocessing is done, then the built model is used to find the final result.

\subsection{Data Collection}
Data collection is the first step of any intrusion detection system. IDS is of two types on the basis of location from which the data is collected: a) network-based IDS, and b) Host-based IDS. In the network based IDS, the data is collected from the network, while data is collected from the host in the host-based IDS.  Our study proposes a network-based IDS to test the proposed approach. Some standard datasets such as: NSL-KDD, KDD-cup99 and Kyoto dataset are chosen as the working data sets. These datasets are already described in  Section \ref{datasets}.
\subsection{Data-preprocessing}
 Generally machine learning classifier requires each instance in the input data as a vector of real number. Thus, to turn the nonnumerical values into numerical values, a pre-processing phase is
required. This phase is common for both training and test dataset. It contains two main stages shown as follows.
 a) Data sample transformation: In this phase, the non-numerical  features are converted to numerical values. Second, third, and fourth features (protocol type, service and flag) of all the three standard datasets (KDD Cup 99, Kyoto, and NSL-KDD)are categorical in nature. 
Specific values are assigned to different samples to convert these features into numerical types such as for protocol type: 'TCP' = 1, 'UDP' = 2 and 'ICMP' = 3, for service type 'aol' =1, 'auth'=2, ;bgp'=3 and so on, and for flag 'oth'=1, 'Rej'=2 and so on. In this way, for each feature, the categories are converted to numerical form. \newline
b) Change of class type from non-numeric to numeric: The classes given in the KDD-cup99 datset, and NSL-kdd dataset are 'normal', 'DOS', 'Probe', 'U2R', and 'R2L'. These are assigned as 1, 2, 3, 4 and
5, respectively. \newline

\subsection{Feature Selection}
In \ref{feature_selection} we have discussed the need and the importance of feature selection. In this phase, we used multi-objective optimization for optimizing different objective functions in order to determine a subset of features. The process of feature selection is discussed later in Section \ref{finding_Feature}. Different feature quality measures like mutual information, pearson correlation coefficient, entropy and information gain are calculated and NSGA-II based feature selection approach is applied to get the Pareto optimal solutions.

\subsection{Model Building}
In this phase we applied different available machine learning classifiers on different Pareto optimal feature subsets found from different models; the models are discussed in Section \ref{study of objective functions}. We used the validation data for finding the best subset of features. The subset on which validation data is giving the best result is chosen for building the final model.

\subsection{Finding results for test data}
After building a model, it is applied on the test data to get the results for an unknown set of data samples. For the test data, data-preprocessing is done. After preprocessing the data, the built model is applied on it to get the final result. In the final result, it is mentioned that, in which class the sample belongs to.
\section{The Proposed Multiobjective Feature Selection Approach} \label{Proposed_Approach}
In this paper, we have devised a filter based feature selection approach, which uses the optimization capabilities of fast and elitist non-dominated sorting genetic algorithm (NSGA-II) \cite{deb2000fast} for optimizing different feature quality measures in order to determine the optimal feature subsets.  This can help in achieving the best classification accuracy for the IDS. Three different feature quality measures are optimized simultaneously to get the Pareto optimal feature subsets. The average dissimilarity of the selected features, the average similarity of the non-selected features, and the average standard deviation of the selected features are optimized simultaneously to identify the optimal feature subsets. After selecting the optimal feature subsets, we have applied different machine learning classifiers such as decision tree, support vector machine, random forest, k-nearest neighbour, Adaboost etc. to check the behaviour of obtained feature subsets on different classifiers. Our proposed approach is divided into two stages. First stage deals with the optimization of feature subsets and the second stage deals with the calculation of classification accuracies using differently available classifiers.  Below, we have discussed the process of feature selection and application of different machine learning classifiers on the selected feature subsets. 
\subsection{Chromosome Representation}

In Genetic algorithm, chromosomes are used to represent a solution as discussed in Section \ref{background}. For feature selection problem,  binary encoding is used. If the feature set contains $n$ number of features, then  a string with $n$ binary
digits is used. Each binary digit represents a feature, value of 1 is used to represent the selection of the feature and 0 is used to represent the rejection of the feature. \par Let us consider a problem of selecting features among 10 features. The encoded string would be a string of 10 binary digits. Let  "0001000110" be the encoded string. It means 4th, 8th and 9th features are selected, and remaining others are rejected.   In this way, we have represented our chromosomes.

\subsection{Different Models for Objective Function Evaluations} \label{study of objective functions}
For evaluating the significance of each objective function, we have created different models using different combinations of mutual information, Pearson correlation coefficient, information gain, standard deviation, and entropy. The feature subset is divided into two mutually exclusive subsets namely selected feature subset (SF) and non-selected feature subset (NSF). Selected feature subsets refer to the set of all features which are selected after optimization (features whose corresponding entries in the chromosome are 1s) and non-selected feature subset  (features whose corresponding entries in the chromosome are 0s)
consist of those features, which are not selected after optimization.

\subsubsection{Model-I}
In the first model, we have used mutual information for measuring the similarity between the features, and standard deviation for checking the  attribute values of the selected features.
The first objective function $F_1$(.) is defined as the average of normalized mutual information between selected features. As the goal of feature selection is to remove the irrelevant features from the feature set, so mutual information between relevant features must be low. Thus, the $F_1$ should be minimized. To avoid the overhead of minimization, we have changed $F_1$ to its reciprocal, so that it could also be maximized. 
\begin{equation} \label{NMI1}
F_1(.)=\sum_{f_i,f_j\epsilon SF,f_i\not=f_j}^{} \frac{2 NMI(f_i,f_j)}{|SF|(|SF-1|)}  
\end{equation}

\begin{equation} \label{NMInew}
F_11(.)=\sum_{f_i,f_j\epsilon SF,f_i\not=f_j}^{} \frac{|SF|(|SF-1|)}{2 NMI(f_i,f_j)} 
\end{equation}
The second objective function $F_2$(.) is defined as the average of normalized mutual information between non-selected features and the nearest selected feature.  It indicates that if features which are described by one of the selected features are removed, then mutual information must be high. Thus, the $F_2$ should be maximized.\newline
\begin{equation}  \label{NMI2}
F_2(.)=\sum_{f_i\epsilon NSF,f_j\epsilon SF, f_j= 1NN(f_i)}^{} \frac{ NMI(f_i,f_j)}{|NSF|} 
\end{equation}
where 1NN ($F_i$) returns the first nearest neighbor of the non selected feature from the selected feature subset. We used euclidean distance to find the distance between two features.\newline
The third objective function $F_3$ is defined as the average of the standard deviations of selected features. Larger the variation of values of a feature, more it will help in finding class labels. Thus, $F_3$ needs to be maximized.
\begin{equation}  \label{SD}
F_3(.)=\sum_{f_i\epsilon SF}^{} \frac{SD(f_i)}{|SF|} 
\end{equation}
High value of mutual information between two variables indicates high redundancy of information in the dataset. For fast and accurate classification, it is necessary to have mutually exclusive and irredundant  feature set.  Standard deviation is a measure, which is used to state how values for a feature are deviated from the average. If the values are spread out, the value of standard deviation is high, otherwise low. These two performance measures describe how irredundant, mutually exclusive and larger the domain feature subset is. Thus functions described in Equations \ref{NMInew}, \ref{NMI2} and \ref{SD} need to be maximized. In Model-I, we have optimized  the above objective functions to extract the optimal subset of features. \par
We have developed two variations of Model-I: i) Model-I(a) - maximizing all equations simultaneously, and ii) Model-I(b)- ignoring the standard deviation of 4th and 5th features while maximizing equation \ref{SD}. Since the standard deviations of $src\_byte$ and $dest\_byte$ are  very large (988217.1009 and 33039.9678, respectively), so we ignored these two features in order to get rid of biased optimization, as we have to maximize the average standard deviation. 

\subsubsection{Model-II}
Entropy measures the impurity of a feature. Larger the entropy of a feature is, larger information it will contain. Thus we have considered entropy in place of standard deviation. We have replaced the standard deviation of Model-I by entropy in order to evaluate the significance of standard deviation and entropy. 
\begin{equation}  \label{entropy}
F_4(.)= \sum_{f_i \epsilon SF}^{}\frac{Entropy(f_i)}{|SF|}
\end{equation}
In Model-II, the first two equations are same as Model-I, which are \ref{NMInew}, and \ref{NMI2}. We have changed the third equation with \ref{entropy}. Thus in Model-II, we will optimize the functions reported in equations \ref{NMInew}, \ref{NMI2}, and \ref{entropy}.

\subsubsection{Model-III}
Information gain (IG) measures the change in entropy after adding an attribute of a feature. Thus it can also be used as a similarity measure for feature subsets. We have replaced mutual information of Model-I with IG, not the standard deviation to develop the Model-III. Using IG, we can have three different objective functions denoted as $F_{IG1}$(.), $F_{IG2}$(.) and $F_3$(.). 
The new objective functions are : 
\begin{equation} \label{equ IG1}
F_{IG1}(.)=\sum_{f_i,f_j\epsilon SF,f_i\not=f_j}^{} \frac{|SF|(|SF-1|)}{2 IG(f_i,f_j)} 
\end{equation}
\begin{equation} \label{equ IG2}
F_{IG2}(.)=\sum_{f_i\epsilon NSF,f_j\epsilon SF, f_j= 1NN(f_i)}^{} \frac{ IG(f_i,f_j)}{|NSF|} 
\end{equation}
The functions described for Model-III in equations \ref{equ IG1}, \ref{equ IG2} and \ref{SD}  need to be maximized, in order to get the optimal feature subsets. \par
Similar to Model-I, we have created two variations of model-III also; Model-III(a) and Model-III(b)- one for optimizing all the objectives simultaneously and other which ignores the std. deviation of 4th and 5th features while maximizing the std. deviation of the selected features.

\begin{table}
\begin{center}
\caption{Description of different models }
\begin{tabular}{|p{0.2\linewidth}|p{0.7\linewidth} |}
\hline Model & Objective functions\\
\hline Model-I & Mutual information and standard deviation  \\
\hline Model-II &  Mutual information and entropy \\
\hline Model-III &  Information gain and standard deviation\\
\hline Model-IV & Information gain and entropy \\
\hline Model-V & Pearson correlation coefficient and standard deviation \\
\hline Model-VI & Pearson correlation coefficient and entropy \\

\hline
\end{tabular}

\label{Models}

\end{center}
\end{table}

\subsubsection{Model-IV}
Like Model-II, we have replaced the standard deviation of Model-III with entropy in order to compare the significance of standard deviation and entropy. Thus for Model-IV, we have to optimize functions reported in Equations \ref{equ IG1}, \ref{equ IG2}, and \ref{entropy}.

\subsubsection{Model-V}
Pearson correlation coefficient (PCC) can be used in place of mutual information. PCC measures the linear correlation between two random features. 
It can also be used as selection criteria for features. If its value between two features is less, they are less correlated, otherwise they are highly correlated.  Thus, PCC value can also be used as an optimization criteria which will measure the similarity between the features subsets. For Model-v, we  have three different objective functions denoted as $F_{p1}$(.), $F_{p2}$(.) and $F_3$(.). We have replaced  mutual information by Pearson correlation coefficient, not the standard deviation. The new objective functions are : 
\begin{equation}   \label{PCC1}
F_{p1}(.)=\sum_{f_i,f_j\epsilon SF,f_i\not=f_j}^{} \frac{|SF|(|SF-1|)}{2 PCC(f_i,f_j)} 
\end{equation}
\begin{equation}   \label{PCC2}
F_{p2}(.)=\sum_{f_i\epsilon NSF,f_j\epsilon SF, f_j= 1NN(f_i)}^{} \frac{ PCC(f_i,f_j)}{|NSF|} 
\end{equation}

In Model-V, the functions mentioned in equations \ref{PCC1}, \ref{PCC2} and \ref{SD} need to be maximized, in order to get the optimal feature subsets using Pearson correlation coefficient and standard deviation.
Similar to Model-I and Model-III, we have created  two variations of model-V also, Model-V(a) and Model-V(b)- one for optimizing all the objectives simultaneously and other which ignores the std. deviation of 4th and 5th feature while maximizing the std. deviation of the selected features.
\subsubsection{Model-VI}
Like Model-II, Model-IV, we have replaced the standard deviation of Model-V too in order to compare the performance. Thus in Model-VI, it requires to optimize the functions listed in equations \ref{PCC1}, \ref{PCC2}, and \ref{entropy}.

\subsection{Finding optimal feature subsets} \label{finding_Feature}
We have utilized the steps of non-dominated sorting genetic algorithm-II, a popular multi-objective genetic algorithm, to obtain optimal feature subsets with respect to different objective functions. We have created different models after optimizing different feature quality measures using NSGA-II as tabulated in Table \ref{Models}. We have optimized different combinations of feature quality measures to develop different models using NSGA-II. It is a fast and elitist way of optimizing different objective functions to obtain a set of Pareto optimal solutions. 

\begin{algorithm}
\caption{Proposed Unsupervised Feature Selection Algorithm}
\label{alg:proposed}

\begin{algorithmic}[1]
\Procedure{Feature\_selection}{}
\State $A  \gets $  Randomly generated initial population
\State $ Mutation\_rate = 0.0244 $,  $ Crossover\_rate = 0.9 $, $ Pop\_size =  100 $
\State $ Max\_generation =  200 $,  $ num\_generation = 0$
\While{$num\_generation \not =\phi $}
	\While{$A\not= \phi$}
	\State calculate  value of each objective function for each individual in A
	\EndWhile
    \While{$A\not= \phi$}
	\State $ Fronts \gets  $ calculate the non-dominated fronts for each individual in A
	\EndWhile
	\While {$ A\not =\phi$}
		\While{$Fronts \not= \phi$}
			\State calculate the crowding distance of each  individual of A
		\EndWhile
	\EndWhile
	\State $B \gets$ []
	\While{$size of B \not=  population\_size $}
		\State  $ a1 \gets $ the fittest individual using the tournament selection
		\State $ a2 \gets $ the fittest individual using the tournament selection 
		\State $ n\_c \gets $ random.uniform(0,1)
 		\If{$ n\_c$  $>$ $Crossover\_rate$}
 			\State $A \gets $ Cross-over of $ a1 $ and $ a2 $ 
		 	\For {Each feature in the individual}
 				\State $ n\_m \gets $ random.uniform(0,1)
          		\If{$ n\_m$  $>$  $ mutation_rate $}
            		\State flip the feature values
                \EndIf
             \EndFor
           	 \State 	Append the new offspring to B  
        \EndIf
	\EndWhile
	
	\algstore{myalg}
\end{algorithmic}
\end{algorithm}

\begin{algorithm}                     
\begin{algorithmic} [1]                   
\algrestore{myalg}

    \State $ C \gets A \cup B$
    \For{ each individual in c}
    	\State calculate front and crowding distance 
    \EndFor
    \State $ A \gets  $ []
    \While {$size of A\not= Population\_size  $}
    	\For{Each fronts}
        	\State Assign the fronts to A on the basis of crowded comparison operator 
         \EndFor
	\EndWhile
  \State  $ num\_generation \gets num\_generation + 1 $
\EndWhile
 \State return A as the set of  optimal feature subsets  
\EndProcedure
\end{algorithmic}
\end{algorithm}

In Algorithm 1, we have discussed the algorithm for obtaining the optimal feature subsets proposed in the current work.  Here first, the population is initialized by randomly generating some solutions. Then we calculate the values of different objective functions for individual solutions in the population. Values of different objective functions are used to identify fronts utilizing non-dominated sorting approach. After that, crowding distance values are calculated for solutions of different fronts. Then we generate another population from the initial population by using selection, cross-over, and mutation operations. Now the initial population and new off-springs are combined together in order to maintain the elitist nature of NSGA-II. On the combined population, fronts and crowding distance values are calculated again. Then using the crowding distance comparison operator, the best individuals are selected for the next generation. This whole process is repeated until the number of generations equals the maximum number of generations.  After completion of the maximum number of generations, a population of solutions is generated which contains the set of optimal feature subsets.

\subsection{Applying machine learning classifiers on optimal subsets}
We have identified optimal feature subsets using NSGA-II after optimizing different feature quality measures. The optimization of these objective functions leads to the identification of mutually exclusive and highly relevant feature subsets. Since NSGA-II is a multiobjective optimization approach, it provides a set of optimal feature subsets on the final Pareto optimal front. We have applied differently available machine learning classifiers such as decision tree, support vector machine, random forest, k-nearest neighbour classifier, Adaboost, multi-layer Perceptron on the obtained optimal feature subsets for developing some intrusion detection systems. 

\section{Simulation results} \label{Results}
We have performed the simulation of our proposed algorithm on KDD Cup 99, NSL-KDD dataset and Kyoto 2006 datasets. In Section \ref{Results_KDD}, we have tabulated all the results for KDD dataset, in section \ref{Result_NSL} we have mentioned the results of NSL-KDD dataset  and in Section \ref{Resultkyoto}, the results for Kyoto datasets are mentioned.

\subsection{Results on KDD-99} \label{Results_KDD}
Our IDS framework works in two phases: first is the identification of optimal feature subsets and second is the development of best IDS classifiers utilizing the best subset of features identified in the first stage.

\subsubsection{Phase 1: Finding optimal feature subsets using NSGA-II} \label{optimal}
For finding Pareto optimal subsets, we have performed simulations for each of the models discussed in Table \ref{Models}. We have tabulated the maximum, minimum and average lengths of feature subsets obtained from all the models in Table \ref{Length_of_feature_subset}.  It is observed   that feature no. 0, 1, 2, 3, 4, 5, 6, 7, 10, 12, 15, 19, 21, 22, 23, 25, 29, 31, 32, 33, 36, and 39 are present in maximum number of models. Only some features are different in different model's best feature subsets. Thus it is clearly demonstrated that  all the obtained models select important features.  Thus it also validates our assumption about different fitness functions.

\begin{table}

\begin{center}
\caption{Lengths of optimal feature subsets obtained after optimizing different models by the MOO based feature selection technique }
\begin{tabular}{|p{0.2\linewidth}|p{0.1\linewidth}|p{0.1\linewidth}|p{0.1\linewidth}|}
\hline Models &   Max. length & Min. length &  Avg. length \\
\hline Model-I(a) & 22 & 12  &  18  \\
\hline Model-I(b)  & 22 & 14  &  19  \\
\hline Model-II &  23 & 15  & 20    \\
\hline  Model-III(a) & 23 & 11  & 20   \\ 
\hline  Model-III(b)  & 22 & 14  & 18  \\ 
\hline  Model-IV & 23  & 16  & 20   \\ 
\hline Model-V(a) & 35  & 23 & 30   \\ 
\hline  Model-V(b)  &  33 & 23  & 28   \\ 
\hline  Model-VI &   32 & 23 & 28   \\ 
\hline
\end{tabular}
\label{Length_of_feature_subset}
\end{center}
\end{table}

\begin{figure}
  \centering
  \includegraphics[width=.7\linewidth]{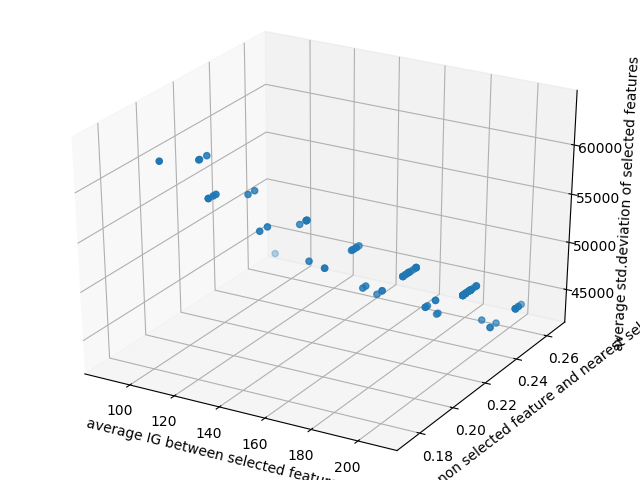}
  \caption{Pareto optimal solutions for Model-III(a)}
  \label{Model_4_plots}
\end{figure}

\subsubsection{Using multi-class classification}
The standard KDD-cup99 dataset comprises of five different classes, so we have opted for multi-class classification. We have applied different classifiers on all the feature subsets obtained from each of the optimizing cases as discussed in section \ref{optimal}. We have applied different classifiers on the identified feature subsets by different models. After analyzing the results of the classification system, we found that feature subsets obtained after executing Model-III(a), provide best results in terms of accuracy on validaion data. The results are tabulated in table \ref{Model_featuresubset}. We performed different tests by changing the classifier. After changing classifiers, we concluded that decision tree is giving the best result, so we have considered decision tree as the classifier for building the IDS. Here we have tabulated the best case result in Table \ref{IG1}. \par
For Model-I, Model-III, and Model-V, we have executed the MOO two times: i) Maximizing all the equations simultaneously, and ii) Maximizing all equations simultaneously, but ignoring the standard deviation of 4th and 5th feature. After analyzing the length and the features present in the feature subsets of both cases, we concluded that
ignoring standard deviation of 4th and 5th feature did not create a significant amount of change in the feature subsets obtained. This variation in the standard deviation is done just to check the performance of the optimizing algorithm. We analyzed that the lengths of obtained feature subsets and the features selected are almost the same and the classification results are also not much different.

\begin{table}

\caption{Best feature subsets and minimum accuracy on 10-fold cross validation data for different models }
\begin{center}
\begin{tabular}{|p{0.18\linewidth}|p{0.5\linewidth}|p{0.12\linewidth}|}
\hline Model  &  Feature subset & Min Accuracy  \\
\hline Model-I(a) & 0, 1, 2, 3, 4, 5, 11, 12, 15, 16, 19, 21, 22, 23, 25, 27, 29, 31, 32, 33, 36 & 99.01 \\
\hline Model-I(b) & 0, 1, 2, 3, 11, 12, 13, 15, 19, 21, 22, 23, 25, 29, 31, 32, 33, 36, 39 & 99.12 \\
\hline Model-II & 0, 1, 4, 5, 6, 7, 10, 12, 13, 15, 17, 18, 19, 21, 22, 23, 25, 26, 29, 31, 32, 33, 35, 36  & 99.36\\
\hline Model-III(a) & 0, 1, 2, 3, 4, 5, 6, 7, 9, 12, 14, 15, 16, 21, 22, 23, 28, 29, 36, 37, 39 & 99.38   \\
\hline Model-III(b) & 0, 1, 2, 3, 4, 7, 8, 9, 10, 12, 13, 15, 21, 22, 23, 28, 29, 31, 32, 36, 37, 39 & 58.93\\
\hline Model-IV & 1, 2, 3, 4, 5, 7, 8, 9, 12, 13, 15, 16, 17, 18, 19, 21, 22, 23, 26, 28, 29, 30, 31, 32, 33, 36, 37, 39 & 58.94 \\
\hline Model-V(a) &  0, 1, 2, 3, 4, 5, 6, 7, 8, 9, 10, 11, 12, 13, 14, 15, 17, 19, 20, 21, 22, 23, 25, 27, 29, 30, 31, 32, 33, 36  & 79.70 \\
\hline Model-V(b) & 0, 2, 3, 4, 6, 7, 8, 10, 11, 13, 14, 15, 16, 17, 19, 20, 21, 22, 23, 25, 27, 29, 30, 31, 32, 33, 36 & 99.37\\
\hline Model-VI & 0, 1, 2, 3, 4, 5, 6, 7, 8, 10, 11, 13, 15, 16, 17, 21, 22, 23, 28, 29, 30, 31, 32, 33, 34, 35, 36, 37, 38, 39 & 58.45 \\
\hline 
\end{tabular}
\end{center}
\label{Model_featuresubset}
\end{table}

\begin{table}
\begin{center}

\caption{Classification accuracy of different classifiers using best Model-III(a)}
\begin{tabular}{|p{0.14\linewidth}|p{0.1\linewidth}|p{0.1\linewidth}|p{0.1\linewidth}|p{0.06\linewidth}|p{0.12\linewidth}|p{0.06\linewidth}|}
\hline Classifiers & Decision tree & Random forest & KNN & MLP & Adaboost  \\
\hline Max. accuracy & 99.38  & 97.56  &  97.59  &  97.61 &  97.38 \\
\hline Avg. accuracy & 96.85 & 97.56  & 96.57 &  94.56  &  21.52 \\
\hline  Min. accuracy & 98.31 &   92.46 & 95.34  &  85.09 &     58.97   \\
\hline
\end{tabular}
\label{IG1}
\end{center}
\end{table}

\begin{figure}
 \centering
 \includegraphics[width=1.0\linewidth]{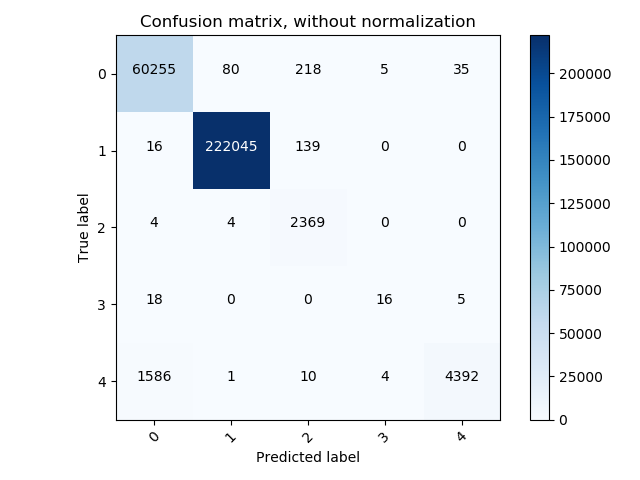}
  \caption{The confusion matrix evaluated from the best feature subset of Model-III(a)}
  \label{fig:cnf_matrix}%
\end{figure}

In fig. \ref{fig:cnf_matrix}, we have shown the confusion matrix for the multi-class classification  of the best subset obtained after the optimization of Model-III(a) for the test data. The classes are not balanced properly and its effect can be clearly observed in the confusion matrix obtained. For classes, normal and Dos, the number of True Positives is very good, both of the classes are classified correctly.  Out of 60593 data samples of the normal class, 601255 samples are classified correctly. For Dos class, out of 222200 samples, 222045 samples are detected correctly.  
The number of samples for U2r attack is very less, only 39. Out of 39 samples, only 16 are classified correctly, while 18 are detected as Dos and 5 as belong to R2l class.  The data samples, which belong to Probe class are classified most accurately. Out of 2377 test samples, 2369 samples are correctly classified, 4 are classified as Dos and 4 as a normal class. 
For R2l class, out of 5993 samples, 4392 samples are classified as R2l, 4 samples as U2r, 1 sample as DOS, and 1566 samples as normal class. In this way, the test data samples are classified efficiently using the classification model.

\begin{table}
\begin{center}
\caption{Overall performance in multi-class classification corresponding to best feature subset obtained by Model-III(a) for KDD dataset }
\begin{tabular}{|p{0.12\linewidth}|p{0.13\linewidth} |p{0.13\linewidth} |p{0.12\linewidth}|p{0.10\linewidth}|p{0.12\linewidth}|}

\hline Class & Accuracy & Detection rate & Precision & False alarm rate & F-measure\\
\hline Normal & 99.32  & 99.44  & 97.37 & 0.7 & 98.39 \\
\hline  Dos &   99.91 & 99.93  & 99.96 & 0.1  & 99.94 \\
\hline Probe & 99.87 & 99.66  & 86.58 & 0.1  &  92.66\\
\hline  U2R & 99.98  &41.02 &  64.00 & 0.003 & 50.00 \\
\hline R2l & 99.43  &  73.28 & 99.09 & 0.01 & 84.25 \\
\hline Weighted average & 99.78 & 99.27 & 99.29 & 0.2  & 99.23 \\

\hline
\end{tabular}
\label{performancemulti1}
\end{center}

\end{table}

In Table \ref{performancemulti1}, the different performance measures like Accuracy, Precision, Detection rate, False alarm rate, F-measure values are reported for the test data.  We have discussed all these metrics in Section \ref{perform}.
Since there are five different classes, so we got five different values of each class for each measure. The weighted average value is considered for giving an unbiased result. The results are discussed in section \ref{results_final} in details.

\subsubsection{Results using binary classification}
The KDD-cup 99 dataset has different classes. Researchers have also done 5-class binary classification (one vs rest) for each class \cite{ambusaidi2016building}, \cite{chebrolu2005feature}. It was done because the classes are not balanced properly, so calculating the performance measures for each class separately may not give accurate results in multi-class classification. We have used binary classification (one vs rest) to show the goodness of the algorithm for a specific class. For each class, we created a new training and test sets. We have assigned the class value as 1 to all the samples belonging to that class and for remaining samples class values are set to  0.  In this way, we got 5 different results for each class. 
We have calculated the accuracy, precision, detection rate, false alarm rate and F-measure values for each of the binary classification results for test data.
In Table \ref{Preformance1}, all the measures are reported, but we have ignored the weighted average section because the classification was binary classification, so there is no need of weighted average.

\begin{table}
\caption{Overall performance in binary-classification for best feature subset of Model-III(a) for KDD dataset}
\begin{center}
\begin{tabular}{|p{0.11\linewidth}|p{0.12\linewidth} |p{0.13\linewidth} |p{0.13\linewidth}|p{0.12\linewidth}|p{0.13\linewidth}|}

\hline Class & Accuracy & Detection rate & Precision & False Alarm Rate & F-measure\\
\hline normal &  98.17  & 92.47  & 99.31 & 0.18 &  95.76 \\
\hline  Dos  & 99.96 & 99.96  &  99.98 &  0.04  & 99.97 \\
\hline Probe  & 99.38 & 58.03 & 88.13  & 0.09  & 69.98 \\
\hline  U2R  &  99.83 & 3.39 & 41.02 &  0.007 & 6.274 \\
\hline R2l & 98.27 &  95.89 & 16.76 &  1.7 & 28.54  \\

\hline
\end{tabular}
\end{center}
\label{Preformance1}
\end{table}

\subsection{Results using NSL-KDD dataset} \label{Result_NSL}
We have divided the  "NSL-KDDTrain+20\%" data into two parts: namely, training and testing. 80\% of data are considered as training data and the remaining 20\% are considered as testing data. The dataset is partitioned in such a manner that, there is no overlap between the training and the test data. \par We have already found a feature subset by the NSGA-II algorithm using KDD Cup dataset. Since, the data distributions for both the NSL and  the KDD datasets are same, so  we do not need to find the feature subset again. We applied different available machine learning classifiers on the NSL-KDD dataset and tabulated the results in Table \ref{performancemulti_NSL}.

\begin{table} 
\begin{center}
\caption{Overall performance in multi-class classification for best feature subset of Model-III(a) for NSL KDD dataset }
\begin{tabular}{|p{0.12\linewidth}|p{0.13\linewidth} |p{0.13\linewidth} |p{0.12\linewidth}|p{0.10\linewidth}|p{0.12\linewidth}|}

\hline Class & Accuracy & Detection rate & Precision & False alarm rate & F-measure\\
\hline Normal & 99.57 & 99.58 & 99.62 & 0.4 &99.60\\
\hline  Dos & 99.90    & 99.93 &99.97 &0.1 & 99.86 \\
\hline Probe  & 99.82 &98.85 &99.13 & 0.08 &98.99 \\
\hline  U2R  & 99.77 & 84.37 &87.09 & 0.1 & 85.71 \\
\hline R2l & 99.92 &66.66 &0.5 & 0.04 & 57.14 \\
\hline Weighted avg.  & 99.83 & 99.16 & 98.73 &0.18 & 98.92\\
\hline
\end{tabular}

\label{performancemulti_NSL}
\end{center}
\end{table}

\begin{table} 
\caption{ Overall performance in multi-class classification  for best feature subset of Model-III(a) for Kyoto dataset}
\begin{center}
\begin{tabular}{|p{0.11\linewidth}|p{0.12\linewidth} |p{0.13\linewidth} |p{0.13\linewidth}|p{0.11\linewidth}|p{0.13\linewidth}|}

\hline Class & Accuracy & Detection rate & Precision & False Alarm Rate & F-measure\\
\hline Known attack &   99.65  &   99.70 & 99.63  & 0.3  &  99.67  \\
\hline  normal &  99.65 & 99.59  & 99.68  &  0.2 &   99.63   \\
\hline unknown attack & 99.99  & 97.5  &  97.08  &  0.004 &  97.29   \\
\hline weighted avg. & 99.65 & 99.65 & 99.65 & 0.3 & 99.65\\

\hline
\end{tabular}
\end{center}
\label{Preformance_Kyoyto}
\end{table}

\subsection{Results using Kyoto 2006+ } \label{Resultkyoto}
We have also used Kyoto 2006+ dataset for evaluating the performance of the proposed feature selection algorithm. In Section \ref{Kyoto}, the dataset and the corresponding attributes are discussed. Some features of Kyoto dataset are also categorical in nature, such as service, flag and protocol. We have applied the data pre-processing part here also. We have chosen the data of 27th August 2009. For testing the performance of the algorithm, we have used 10 fold cross-validation method. We applied NSGA-II based feature selection technique with the proposed models (Section \ref{study of objective functions}) to obtain a set of feature subsets which are optimal with respect to the objective functions of different models. After finding the subsets, we applied 10 fold cross validation to get the best results. As we have found that Decsion Tree attains the best result for other datasets, so we applied decision tree for finding the results. The results are shown in Table \ref{Preformance_Kyoyto}.

\section{Final results and Comparison of our proposed method with other classifiers} \label{results_final}
We have proposed a new unsupervised feature selection approach using multi-objective optimization. In this section, we have discussed the final results of multi-class classification and the binary classification also. Since we have considered different models for optimization, so it would be good to discuss the final results separately. Our algorithm has performed well in all the cases, but in this section, we have reported only the best cases.  In Table \ref{performancemulti1} , we have reported different performance measures for different classes. Our proposed approach attained an accuracy of 99.32\% for normal, 99.91\% for Dos, 99.87\% for probe, 99.98\% for U2R and 99.43\% for R2l classes. After calculating the weighted average of accuracies, our proposed system attained overall 99.78\% accuracy.  We have to detect the attack classes, so we found the detection rates for different classes separately. Our proposed system attained a detection rate  of 99.44\% for normal, 99.93\% for Dos, 99.66\% for probe, 41.02\% for U2R and 71.01\% for R2l classes. The number of samples in Probe class is very less, so we attained a very less detection rate for this case.  
We have considered weighted average for calculating the overall performance of the system. The reason for taking weighted average is that the classes are not balanced. We have already tabulated the distribution of instances in Table \ref{Data}. As we can see that, for classes like R2l, U2R and probe, comparatively sample instances are very less. Thus it is a good choice to evaluate weighted average. With weighted average, proposed system attained an accuracy of 99.78\%, detection rate of   99.27\%,  a precision of 99.29 \%, a false alarm rate of 0.2\% and  a F-measure of 99.23\% . \par In order to show that performance improvements obtained by our proposed approach are not happened by chance but those are statistically significant, we have performed statistical t-test over results obtained by the proposed MOO-based approach with different models and existing best model (LSTM based \cite{kim2017effective}). We have reported the results of statistical significance tests in Table \ref{models_test}. From this table, it is clear that all the models proposed in this paper  attain improved results over the LSTM based approach \cite{kim2017effective}. LSTM model requires a huge amount of labelled data for learning architecture and weight values whereas our proposed unsupervised feature selection approach requires no labeled data while selecting features. Labeled data with limited feature set is used only in developing decision tree based classifier. \par We have also tabulated the results of binary classification of KDD-dataset in table \ref{Preformance1}. We got an accuracy of 98.17\%, 99.96\%, 99.38\%, 99.83\%, and 98.27\%  for normal, Dos, Probe, U2R, and R2l class respectively. We have also tabulated other performance measures in the table too. \par For 20\% of NSL-KDD dataset, we got an overall accuracy of 99.83\%, 99.16\% detection rate, 98.73\% precision, 0.18\% of false alarm rate, and 98.92\% of F-measure. These results are tabulated in \ref{performancemulti_NSL}. We got an overall accuracy of 99.65\%, detection rate of 99.65\%, 0.3\% of false alarm rate and 99.65\% of F-measure for the Kyoto dataset also. These results are tabulated in table \ref{Preformance_Kyoyto}.
 
\begin{table}
\begin{center}
\caption{t-test and p-test values for different models}
\label{models_test}
\begin{tabular}{|p{0.2\linewidth}|p{0.15\linewidth} |p{0.2\linewidth} |p{0.2\linewidth}|}
\hline Models & t-value & Is P-value $>$ than 0.00001 & Is significant at P $>$ than 0.5 \\
\hline Model-I(a) & -22.17 & Yes & Yes \\
\hline Model-I(b) & -46.864 & Yes & Yes \\
\hline Model-II & -64.25 & Yes & Yes \\
\hline Model-III(a) & -10.14 &  Yes & Yes \\
\hline Model-III(b) & -8.14279 & Yes & Yes \\
\hline Model-IV & -10.93 & Yes & Yes \\
\hline Model-V(a) &  -11.28 & Yes & Yes \\
\hline Model-V(b) & -12.735 & Yes  & Yes  \\
\hline Model-VI & -49.72 & Yes & Yes \\
\hline
\end{tabular}
\end{center}
\end{table}

\par  In Table \ref{Table_comp}, we have presented a tabular comparison of results of different published works with respect to accuracy values and the achieved accuracy by our proposed model. The best accuracy and detection rate for the current dataset are achieved by Ref. \cite{kim2017effective}. Authors had applied long short term memory recurrent neural network with all the 41 features. They got 97.54 \% accuracy and 98.95 \% detection rate. Our proposed model gives a set of feature subsets in which the maximum length of the feature subset is 23, minimum length is 11, and the average length of all feature subsets is 20. Thus it can be said that our proposed classifier will be less costly as compared to the LSTM based classifier, as it uses all the 41 features for the classification purpose because our classifier uses only 21 features.


\begin{table} 
\begin{center}
\caption{Comparison of our proposed method with other classifiers performing multi-class classification for KDD-Cup dataset}
\label{Table_comp}
\begin{tabular}{|p{0.17\linewidth}|p{0.11\linewidth} |p{0.13\linewidth} |p{0.13\linewidth}| p{0.1\linewidth}|}
\hline classifier & Precision & Detection rate & Accuracy & False alarm rate\\
\hline \cite{devaraju2014performance} GNNN  & 87.08 & 59.12 & 93.05 & 12.46 \\ 
\hline  \cite{kim2016long} LSTM &   & 98.88 & 96.93 & 10.04  \\
\hline \cite{wang2010new}fuzzy clustering & & & 96.75 & \\
\hline \cite{kim2017effective} LSTM    & 97.69 & 98.95 & 97.54 & 9.98\\
\hline \bf{Proposed Algorithm} & \textbf{99.78} & \textbf{99.29}   & \textbf{99.27} & \textbf{0.2}\\
\hline
\end{tabular}
\end{center}
\end{table}

We have also compared our method with other feature selection based Intrusion detection system. In table \ref{Table_feature_comp}, we have tabulated some other works, in which authors have selected optimal feature subsets in order to achieve good accuracy.  In \cite{aghdam2016feature}, authors observed the accuracy of 98.9 \% using ant colony optimization strategy. At the same time, our system got an overall accuracy of 99.78\%.  We have not compared our performance metrics with \cite{ambusaidi2016building} because they have used classes for detecting relevant set of feature subset. They got an accuracy of 99.79\%, detection rate of 99.46\%, and false alarm rate of 0.13\% which is not very much high in compare to our result.  \par
We have also achieved an accuracy of 99.57\%, 99.90\%, 99.82\%, 99.77\%, and 99.92\% for Normal, DOS, Probe, U2R, and R2L class respectively for the test dataset of NSL-KDD dataset which is better that the previous best result described in \cite{aljawarneh2018anomaly}. They got an accuracy of 99.7\%, 99.9\%, 96.2\%, 99.1\%, and 97.9\% for normal, Dos, Probe, U2R, and R2l respectively. Thus it can also be seen that our method is performing well on NSL-KDD dataset also.

\begin{table}

\caption{Comparison with other feature selection algorithms for KDD-Cup dataset}
\label{Table_feature_comp}
\begin{center}
\begin{tabular}{|p{0.5\linewidth}| p{0.2\linewidth}|}
\hline Method & Accuracy\\
\hline Ant colony optimization method \cite{aghdam2016feature} & 98.9 \\
\hline Cuttlefish optimization \cite{eesa2015novel} & 91.98 \\
\hline Chi-square based \cite{thaseen2017intrusion} & 98 \\
\hline clustering feature \cite{horng2011novel} & 95.7\\
\hline KDD-winner \cite{pfahringer2000winning} & 91.8\\
\hline KDD-runner up \cite{levin2000kdd} & 91.5\\
\hline \textbf{Proposed Method} & \textbf{99.78}\\
\hline

\end{tabular}

\end{center}
\end{table}

\section{Conclusion and future Work} \label{Conclusion and future work}
In this research work, we have devised a new filter-based feature selection technique in multiobjective optimization framework for detecting relevant features from an unlabeled data set. Our goal was to devise a new algorithm for selecting features without using any class label, and without compromising the accuracy of IDS. Originally, there were 41 attributes in the standard KDD-99 dataset, but our designed model uses a maximum of 23 features, an average of 20 and a minimum of 11 features for classification.  Our system gives an accuracy of 99.38\% for multi-class classification. This accuracy is calculated according to equation \ref{accuracy}. 
Since the classes are not balanced properly so finding the weighted average (equation \ref{wt_accuracy} ) 
of accuracy values would be a better choice.  Our system attains a weighted average of 99.78\% accuracy with Decision Tree classifier. To the best of our knowledge, this is the best-reported accuracy compared to state-of-the-art techniques.  In \cite{kim2017effective}, authors have reported an accuracy of 97.54\% and in \cite{ambusaidi2016building} authors have reported an accuracy of 99.80 \%, but here 5-class binary classification problem is solved and class labels are utilized for selecting a suitable set of features. Our devised method attains the best accuracy as well as it is an efficient unsupervised method for selecting relevant features from a huge and unlabeled dataset. We are also getting an weighted average accuracy of 99.83\% for NSL-KDD dataset and 99.65\% for Kyoto dataset. \par In future, we aim to implement some deep learning approaches to classify the system using the feature sets obtained by our proposed feature selection technique. In future, we have also planned to devise some novel MOO-based wrapper algorithms, which can find optimal feature subsets using  Convolution neural network, Autoencoder, recurrent neural network, long short-term neural network for classification purpose.

\bibliographystyle{model1-num-names}
\bibliography{refer}







\end{document}